# Современные направления развития и области приложения теории Демпстера-Шафера (обзор)*

В.К. Иванов[I], Н.В. Виноградова[I], Б.В. Палюх[I], А.Н. Сотников[II]

[I] Тверской государственный технический университет, Тверь, Россия
[II] Межведомственный суперкомпьютерный центр РАН, Москва, Россия

**Аннотация.** В статье представлен обзор публикаций, посвященных современным направлениям развития теории Демпстера-Шафера и ее приложений для различных областей, науки, техники и технологий. Рассмотрены те направления исследований, результаты которых известны не только в узкопрофессиональном научном сообществе, но также доступны широкому кругу потенциальных разработчиков перспективных технических решений и технологий. Показано применение результатов теории в некоторых важных областях человеческой деятельности, таких как производственные системы, диагностика технологических процессов, материалов и изделий, строительство, управление качеством продукции, социально-экономические системы. Основное внимание уделено современному состоянию исследований в рассматриваемых областях, в связи с чем, отбирались и анализировались работы, изданные, как правило, в последние годы и представляющие достижения современных исследований теории Демпстера-Шафера и применения ее результатов.

**Ключевые слова:** вероятность, комбинирование свидетельств, моделирование конфликтов, принятие решений, теория свидетельств, степень правдоподобия, теория Демпстера-Шафера, функция доверия, учет неопределенностей.

## Введение

Математическая теория Демпстера-Шафера (Dempster-Shafer theory–DST), известная также как теория свидетельств, теория очевидностей или теория функций доверия, дает математическую основу для вычисления вероятности события после комбинирования отдельных частей исходной информации об этом событии, полученных из разных источников.

С момента появления первых работ [1, 2] в 1967 и 1976 году соответственно, в мире изданы тысячи трудов по тематике DST, в которых уточняются, развиваются и обобщаются ее базовые положения и концептуальные основы, совершенствуются алгоритмы, описываются новые области применения, выявляются и преодолеваются недостатки теории.

В статье представлен обзор публикаций, касающихся современных направлений развития DST и ее приложений. Учитывая обширность и многоплановость анализируемой тематики, прежде всего, принимались во внимание следующие цели, позволившие сконцентрироваться на конкретных аспектах DST.

– Показать направления современных исследований DST, результаты которых известны не только научному сообществу, но также востребованы и доступны широкому кругу разработчиков перспективных технологий. При этом значительное внимание уделено современному состоянию применения DST в различных областях теории принятия решений. Это предмет первой части обзора, включает вопросы комбинирования свидетельств, моделирования конфликтов, применение функций доверия в экспертных системах и онтологиях.

– Показать применение результатов DST в различных областях человеческой деятельности, а именно в производственных системах, диагностике дефектов материалов и изделий, системах принятия решений в строительстве, на транспорте, в управлении качеством продукции, в естественнонаучных и социально-экономических исследованиях. Этому посвящена вторая часть обзора.

Следует отметить, что, начиная с 2005 г. и по настоящее время, наблюдается рост числа публикаций, посвященных различным аспектам теории принятия решений. В связи с этим авторами были рассмотрены основные результаты этого периода, полученные российскими и зарубежными учеными, и которые в настоящее время формируют тренды развития и применения DST. Работы, включенные в обзор, представляют, в основном, примеры направлений исследования и применения DST, а также анализ полученных результатов. Более полную информацию для каждой рассмотренной области исследований и прикладных разработок можно найти в приведенном списке публикаций.

Подбор материалов для обзора осуществлялся авторами на основе многолетнего собственного опыта исследования DST и использования ее результатов для решения прикладных задач, а также с помощью поисковых систем Интернета и в специализированных базах данных (таких как ieeexplore.ieee.org и dl.acm.org). Всего было проанализировано около пятисот источников, относящихся к рассматриваемой тематике.

# 1. Основные положения теории Демпстера-Шафера

## 1.1. Концептуальные основы

Рассмотрим кратко основные положения DST [2]. Пусть имеется множество объектов и множество всех возможных свойств, которыми они могут обладать. Необходимо по известному набору признаков определить объекты, обладающие заданными свойствами. Основная идея DST заключается в том, что некоторая мера вероятности может быть отнесена не только к отдельным элементам множества событий в предметной области, но и в целом к некоторому подмножеству этого множества. Причем более подробное распределение этой частичной меры вероятности по подмножеству неизвестно. Одновременно могут быть неизвестны и некоторые меры вероятности, относящиеся непосредственно к элементам множества событий. При этом закрепление общей меры вероятности за подмножествами множества объектов с заданными свойствами может происходить по следующим причинам. Во-первых, из-за неоднозначности решения задачи экспертной классификации, когда эксперт не может точно определить уровень значимости признака для одного, отдельно взятого объекта. Во-вторых, в случае неточности результатов определения значения самого признака. В-третьих, при назначении общего уровня доверия для всех признаков.

Степень уверенности в наличии заданного свойства у объектов выражается функцией доверия:

$$Bel(A_i) = \sum_{A_j \subseteq A_i} m(A_j), \qquad (1)$$

где $A_i$ - событие, соответствующее наличию заданного свойства у объекта или некоторого множества объектов; $A_i \subseteq C$; $C$ – полное множество событий, которое является результатом решения задачи; $m(A_j)$ – базовое распределение вероятностей события $A_j$; $m(A_j) \in [0,1]$. $Bel(A_i)$ обладает следующими свойствами: $Bel(\emptyset) = 0$, $Bel(A_i) \in [0,1]$ и $Bel(C) = 1$. Степень правдоподобия наличия заданного свойства у объектов выражается функцией:

$$Pl(A_i) = 1 - Bel(\overline{A_i}) = 1 - \sum_{A_i \cap A_j = \emptyset} m(A_j). \qquad (2)$$

По аналогии с теорией вероятности величины $Bel(A_i)$ и $Pl(A_i)$ можно рассматривать как нижнюю и верхнюю границы вероятности наличия заданного свойства объектов, принадлежащих подмножеству $A_i \subseteq C$, предполагая существование некоторой истинной вероятности $p(A_i)$:

$$Bel(A_i) \le p(A_i) \le Pl(A_i). \qquad (3)$$

В соответствии с правилом Демпстера объединение различных свидетельств (событий) с распределениями вероятностей $m_1$ и $m_2$ выполняется следующим образом:

$$m(C_N) = \frac{1}{1 - m(\emptyset)} * \sum_{A_i \cap A_j \neq \emptyset} m_1(A_i) * m_2(A_j), \qquad (4)$$

где $m(C_N)$ - функция объединения событий $A_i$ и $A_j$, соответствующих наличию у объекта свойств $C_N$; $m(\emptyset) = \sum_{A_i \cap A_j = \emptyset} m_1(A_i) * m_2(A_j)$.

Правило Демпстера ассоциативно и коммутативно, что позволяет объединять аналогичным образом и большее число свидетельств.

Решение, найденное с помощью DST, представляет собой полную группу событий, что, с одной стороны, позволяет лучше оценить реальную обстановку, а с другой стороны, дает возможность дальнейшего уточнения решения задачи при появлении дополнительных признаков. Увеличение количества признаков позволяет проводить более точное решение задачи с локализацией заданного свойства до уровня одного-двух объектов.

## 1.2. Пример использования

Для иллюстрации возможностей DST рассмотрим решение задачи из области разработки диагностических систем для управления качеством продукции в многостадийном производстве [3]. Основным условием нормального функционирования такого производства является отсутствие кризисных состояний и ухудшений показателей эффективности $X_k = \left[\underline{x_k}, \overline{x_k}\right]$ у отдельных агентов $A_i \in A_k$. Признаком кризиса является выход значения $X_k$ за допустимые нормативные значения $D_k = \left[\underline{d_k}, \overline{d_k}\right]$. Существует два типа ошибок, возникающих при проверке диагностических гипотез: ошибки 1-го рода или «ложная тревога» и ошибки 2-го рода или «отсутствие тревоги при ухудшении показателей эффективности».

Пусть $A^*$ - множество кризисных агентов. Тогда диагностическая процедура есть поиск решений следующего логического уравнения:

$$F: P^* \to A^*, \qquad (5)$$

где $F$ – предикатная функция вида $F = \wedge(\vee A_i)$, $P_k = 1$ и $P^* = \{P_k | P_k = 1\}$ – множество зарегистрированных ухудшений показателей эффективности; $P_k$ – значение индикаторной функции.

Возникновение любого диагностируемого кризисного состояния $s \in S$ агента $A_i$, характеризуется необходимым и достаточным условиями:

$(\forall s \in S)(P_k = 0) \leftrightarrow (A_k \cap A^* = \emptyset)$ – необходимое условие; $\qquad (6)$

$(\forall s \in S)(P_k = 1) \leftrightarrow (A_k \cap A^* = A^*)$ – достаточное условие. $\qquad (7)$

Отметим, что невыполнение условия (6) приводит к ошибкам 2-го рода, а условия (7) – к ошибкам 1-го рода. Покажем, как методы DST позволяют снизить требования условий (6) и (7) до вида:

$$(\forall s \in S)(P_k = 0) \leftrightarrow P^* \neq \emptyset. \qquad (8)$$

Предположим, что степень влияния $X_k$ на кризисное состояние агента $A_i$ оценивают эксперты числом в диапазоне [0; 100]. Пример такой оценки представлен в Табл. 1.

Табл.1. Результаты экспертной оценки

|       | $A_1$ | $A_2$ | $A_3$ | $A_4$ | $A_5$ |
|-------|-------|-------|-------|-------|-------|
| $X_1$ | 20    | 40    |       |       | 60    |
| $X_2$ | 15    |       | 40    |       |       |
| $X_3$ | 90    | 50    | 30    | 30    |       |

Пусть $X_1 = [42{,}8\%; 52{,}4\%]$ и $X_3 = [53{,}0\%; 58{,}0\%]$ с нормативным интервалом $S = [50{,}55\%]$. Индикаторная функция $P_k$ при выходе интервала $X_k$ за верхнюю границу $\overline{d_k}$ имеет вид:

$$P_k = \begin{cases} 0, \text{если } \overline{x_k} \leq \overline{d_k} \\ 1, \text{если } \underline{x_k} \geq \overline{d_k} \\ \dfrac{\overline{x_k} - \overline{d_k}}{\overline{x_k} - \underline{x_k}}, \text{если } \underline{x_k} < \overline{d_k} < \overline{x_k} \end{cases} \qquad (9)$$

А при выходе интервала $X_k$ за нижнюю границу $\underline{d_k}$ индикаторная функция $P_k$ имеет вид:

$$P_k = \begin{cases} 0, \text{если } \underline{x_k} \geq \underline{d_k} \\ 1, \text{если } \overline{x_k} \leq \underline{d_k} \\ \dfrac{\underline{d_k} - \underline{x_k}}{\overline{x_k} - \underline{x_k}}, \text{если } \underline{x_k} < \underline{d_k} < \overline{x_k} \end{cases} \qquad (10)$$

В соответствии с (10) для показателя $X_1$ вероятность его ухудшения $P_{x1} = 0{,}75$. Для показателя $X_3$ в соответствии с (9) вероятность его ухудшения $P_{x3} = 0{,}6$.

Из Табл.1 имеем множества экспертных оценок кризисных состояний $A_{x1}$ и $A_{x3}$:

$$A_{x1} = \{(a_1; 0{,}2), (a_2; 0{,}4), (a_5; 0{,}6)\}$$

и $A_{x3} = \{(a_1; 0{,}9), (a_2; 0{,}5), (a_3; 0{,}3), (a_4; 0{,}3)\}$.

Нормированные распределения вероятностей кризисных состояний ($\sum m_{xi} = 1$) следующие:

$m_{x1} = \langle a_1; a_2; a_5 \rangle = \langle 0{,}17; 0{,}33; 0{,}50 \rangle$ и $m_{x3} = \langle a_1; a_2; (a_3 \vee a_4) \rangle = \langle 0{,}55; 0{,}30; 0{,}15 \rangle$.

Перераспределенные распределения вероятностей $m'_x = m_x * P_k$ и $m'_x(A) = 1 - P_k$, представлены ниже:

$m'_{x1} = \langle a_1; a_2; a_5; A \rangle = \langle 0,13; 0,25; 0,48; 0,25 \rangle$

$m'_{x3} = \langle a_1; a_2; (a_3 \vee a_4); A \rangle =$
$= \langle 0,33; 0,18; 0,09; 0,40 \rangle$.

Объединение $m_{xi}$ осуществляется на основе правила Демпстера (4). В соответствии с ним $m(\emptyset) = 0,35$. Для гипотез о кризисных агентах:

$$m(a_1) = \frac{(m_{x1}'(a_1) * m_{x3}'(a_1) + m_{x1}'(A) * m_{x3}'(a_1) + m_{x1}'(a_1) * m_{x3}'(A))}{m(\emptyset)}$$

$= (0,13 * 0,33 + 0,25 * 0,33 + 0,13 * 0,40)/0,65 = 0,26;$  $m(a_2) = 0,28;$  $m(a_3 \vee a_4) = 0,03;$ $m(a_5) = 0,29; m(A) = 0,15.$

Итоговое решение: $a_1[0,26; 0,41]$; $a_2[0,28; 0,43]$; $a_3[0,03; 0,18]$; $a_4[0,03; 0,18]$; $a_5[0,29; 0,44]$; $A[1,0; 1,0]$.

Таким образом, мы оставляем гипотезу о кризисном состоянии агента $A_5$ для дальнейшей проверки, в результате которой может быть установлена причина кризиса. Если бы анализ проводился с использованием традиционных методов с использованием нечетких множеств, эта гипотеза была бы отвергнута, что привело бы к ошибке 2-го рода.

## 2. Развитие DST и ее приложений

В книге [4] собраны классические работы по DST, которые составляют ее исчерпывающую фундаментальную основу. В ней представлены результаты исследований ведущих специалистов в области DST и ее приложений: A. Chateauneuf, A. Tversky, A.P. Dempster, D. Dubois, E.Y. Shortliffe, E.H. Ruspini, G. Rogova, G. Shafer, H. Prade, H.T. Nguyen, J. Gordon, J.-Y. Jaffray, J.A. Barnett, J. Yen, J. Kohlas, L. Liu, J.D. Lowrance, N.L.Zhang, Ph. Smets, P.P. Shenoy, R.P. Srivastava, R. Logan, R.R. Yager, T. Denœux, Th.D. Garvey, Th.M. Strat. В этих работах отражены практически все актуальные аспекты DST: методы рассуждений, обобщение байесовского вывода, функции доверия, распределение вероятностей, вычислительные методы, информационная энтропия, разрешение конфликтов между свидетельствами и др.

Отметим здесь проводящуюся раз в два года конференцию BELIEF, организованную международным обществом по функциям доверия и их применению Belief Functions and Applica-tions Society (BFAS). BFAS (www.bfasociety.org) было создано для содействия исследованиям и применению в различных областях функций доверия и их расширений. Также в деятельности BFAS большое значение придается использованию функций доверия и DST в целом с другими теориями и методами. Первая конференция BELIEF прошла в 2010 г. в Бресте (Франция), а последняя – в 2016 г., в Праге (Чешская Республика), материалы которой представлены в [5]. В статьях этого сборника описываются последние разработки теоретических вопросов и приложений DST в различных аспектах. Среди них правила комбинирования свидетельств, управление конфликтами между свидетельствами, обобщения теории информации, обработка изображений, применение методов DST в материаловедении и навигационных системах.

В [6] отмечается, что в последние годы количество статей по теории и практике применения DST переживает бум (более 1200 опубликованных материалов только в 2014 году). Наблюдается явный рост таких публикаций, особенно в восточноазиатском регионе. Области применения результатов исследований и разработок также разнообразны: от геоинформационных до диагностических систем. Причина этого достаточно очевидна – DST и, в частности, функции доверия являются естественным инструментом в системах принятия решения по многим критериям в условиях высокой неопределенности и недостатка исходных данных, отсутствия подтверждения достоверности данных, использования крайне редких значимых событий. Тематика и направления исследований исключительно широки. В них входят: развитие понятия функции доверия [7], схемы их агрегирования [8], анализ числовых характеристик [9], решения для слабо доверительных ситуаций [10], использование эмпирических моделей [11], специфические расширения теории вероятностей [12].

Отдельно следует сказать об использовании в исследованиях, связанных с DST, фундаментального понятия информационной энтропии. В [13] предлагается новое определение энтропии при оценках распределения вероятностей, используемых в функциях доверия. Сформулировано пять желаемых свойств энтропии, адаптированной к DST, которые основываются на шенноновском понимании энтропии. Обсуж-даются некоторые открытые вопросы, в частности, удовлетворение предложенного определения энтропии свойству субаддитивности. В [14] определяются семейства мер неопределенности, связанных с распределением вероятностей, и предлагается соответствующая формулировка энтропии, а в [15] исследуются базовые свойства рассматриваемых мер энтропии.

## 2.1. Комбинирование свидетельств и моделирование конфликтов

Важным аспектом DST является разработка и исследование правил комбинирования свидетельств (доказательств), полученных из нескольких источников. При этом особое значение придается моделированию конфликтующих свидетельств. Ниже представлен ряд работ, в которых подходы к построению правил комбинирования свидетельств, расширяющие возможности DST.

В статье [16] детально анализируются идеи комбинирования свидетельств, заложенные в DST, с акцентом на недостаточность, в ряде случаев, известных подходов. Авторы берут за основу правило комбинирования Ягера [17] и, принимая во внимание принцип «меньшинство подчиняется большинству», вводят новое эффективное правило комбинирования свидетельств.

Главной проблемой, рассматриваемой в работе [18], является построение модели, которая при комбинировании свидетельств (знаний) поддерживала бы представление уровня незнания и степени конфликтности между свидетельствами. Работа содержит описание нового правила комбинирования свидетельств и ряда сопутствующих понятий, связанных с представлением уровня неопределенности генерируемых решений.

В работе [19] вводится и исследуется функционал уменьшения незнания после комбинирования ряда свидетельств (индекс уменьшения незнания), полученных из разных источников информации. Также аксиоматически вводится мера конфликта свидетельств, позволяющая количественно оценить степень конфликтности. Статья иллюстрирует достижения российских исследователей в рассматриваемой области.

Обобщенная теория свидетельств предлагается в [20], где исследуются особенности конфликтных ситуаций и предлагаются пути их разрешения, выходящие за рамки классической DST.

В [15] описывается новый метод комбинирования взвешенных высоко конфликтных свидетельств, основанный на определении расстояния между свидетельствами и функциями энтропии свидетельств. Метод позволяет эффективно сблизить разнородные свидетельства, полученные из разных источников.

## 2.2. Экспертные системы и онтологии

Нельзя обойти вниманием такую область применения DST, как экспертные системы. Отметим два исследования, которые отражают некоторые актуальные направления в развитии технологий построения экспертных систем.

В статье [21] описываются проблемы, с которыми может столкнуться разработчик баз знаний. Одна из них – это условия устойчивости взаимосвязанных мнений экспертов, которые используются в правилах вывода. Предлагается проект приложения, использующего DST в классических (основанных на MYCIN-подобных правилах) экспертных системах. Выполнен анализ данного подхода и сформулированы некоторые открытые проблемы.

В работе [22] рассматривается оригинальный подход к решению известной проблемы комбинирования свидетельств, полученных из разных источников. Для решения предлагается дисконтирование свидетельств, основанное на сравнении их векторных представлений. Для вычисления коэффициента дисконтирования используется косинусная мера сходства векторов. Затем дисконтированные свидетельства объединяются правилом Демпстера для получения окончательного результата. На примере показывается эффективность применения подхода в экспертных системах.

Построение онтологий или семантических описаний в различных предметных областях также продолжает оставаться актуальным направлением развития семантических технологий. Эти технологии предлагают мощные средства для представления знаний, а также расширенные возможности анализа рассуждений (правил вывода) для современных приложений. Тем не менее, вопрос адаптации к анализируемому контексту неопределенного представления о реальном мире есть один из ключевых вопросов при построении онтологий. Отсутствие учета неопределенности фактов предметной области рассматривается как серьезный недостаток.

В работе [23] эту проблему исследуют средствами DST. Предлагается подход, при котором на базе DST автоматически создается согласованная структура для применения классических комбинаций информационных единиц в процессе принятия решений. Исследуется ситуация использования данного подхода в иерархической структуре с учетом гипотез, аксиоматических ограничений и свойств объектов онтологий.

Фундаментальный характер онтологий подразумевает, что они описывают только как подтвержденные и достоверные факты предметной области. В реальности для обеспечения этого условия существуют различные вероятностные, нечеткие, либо очевидностные, подходы. В статье [24] собраны и описаны самые популярные инструменты для работы с онтологиями. Авторы дополняют список онтологией DST, которая может быть импортирована в онтологию исследуемой предметной области. Такой подход позволяет имплементировать онтологии с учетом неопределенности собираемых, классифицируемых и систематизируемых фактов.

### 2.3. Программное обеспечение

Несмотря на определенную особенность использования DST в каждом конкретном случае существует очевидная необходимость в достаточно универсальной программной реализации, по крайней мере, обобщенной модели вычислений. Примером такой программной реализации может служить программная библиотека py_dempster_shafer [25], модули которой написаны на языке Python. Средствами библиотеки обеспечивается поддержка нормализованных и ненормализованных функций доверия, различные методы (например, Монте-Карло) для комбинации функций доверия, использование методов для обобщения теоремы Байеса, поддержка различных мер неопределенности, методы конструирования функций доверия, набор примеров использования.

Другой пример [26] касается специализированного программного обеспечения контроллера дискретных событий, который является составной частью диагностической системы с многоступенчатым выбором решений. Реализуемая схема принятия решений основана на новом методе диспетчеризации, обслуживания рабочих мест и ресурсов с использованием сетей Петри и DST. Приведены примеры применения и реализации указанного метода для принятия решений в интеллектуальной диагностике.

Отметим также специальное программное обеспечение, используемое при работе с шаблоном, созданным на основе DST, для его возможного импорта в какую-либо онтологию предметной области с целью учета неопределенности ее семантики [24].

### 2.4. Области приложения DST

Ниже представлены некоторые сведения о применении DST для решения задач принятия решений. Безусловно, список источников, куда входят и работы российских ученых, можно расширить, но основные тренды разработки приложений на основе DST уже определены достаточно ясно.

**Производственные системы.** Актуальность разработки и применения методов и алгоритмов многокритериального принятия решений с использованием DST в производственных задачах очевидна [27]. В качестве примеров приведем следующие направления исследований:
– разработка моделей поведения сложных технических систем [28];
– робототехника [29];
– обеспечение надежности производственных систем, включая разработку новейших методов анализа данных от измерительных устройств [30].

**Анализ экспертных оценок** возможных решений многокритериальных задач - это одна из самых обширных областей применения DST. Довольно большое количество работ посвящено этой проблематике. Выделим здесь некоторые показательные направления:
– анализ групповых экспертных оценок в конфликтных ситуациях [31];
– обработка экспертных субъективных оценок в естественнонаучных сферах, например, при проведении гидрологических исследований [32];
– исследование проблем интеграции лингвистической информации в групповых экспертных оценках [33];
– экспертные системы для информационной поддержки операторов автоматизированных производств [34].

**Диагностические системы в управлении качеством продукции.** Диагностика технических объектов и систем в управлении качеством продукции является еще одним фунда-ментальным направлением в прикладном использовании DST. Отметим многоплановую работу [35], в которой в контексте DST описываются различные аспекты диагностики технических систем и дается подробное описание применения диагностических систем, в том числе, в машиностроении с акцентом на трибологические процессы, в строительстве мостов, диагностике трубопроводов, электростанций, железнодорожных объектов, а также применения диагностических методов в искусстве и культуре. Также заслуживает внимания работа [36], посвященная применению методов неразрушающего контроля (Nondestructive testing – NDT) при обнаружении дефектов в материалах и конструкциях применительно к различным областям от аэрокосмических систем до медицины. Книга позиционируется авторами не только как прагматическое введение в NDT, но и как базис для проведения теоретических исследований, а также использование DST.

Приведем ряд конкретных направлений исследований и разработок на основе DST в области диагностических систем:

– диагностика отказов непрерывных химико-технологических процессов [37,38];

– анализ дефектов сварных швов [39];

– ультразвуковой и вихретоковый контроль изделий в машиностроении, энергетике, самолетостроении, нефтехимии [40];

– контроль качества водораспределительных сетей [41,42];

– диагностика высоковольтных прерывателей [43];

– диагностика двигателей [44,45];

– диагностика сбоев функциональных блоков и компонентов на основе анализа потоков данных [46];

– диагностические системы с многоступенчатым выбором решений [26].

**Обработка изображений в технологиях дистанционного зондирования.** Аэрокосмическое зондирование, имеющее важное экономическое и военное значение, является методом исследований, результаты применения которого используются практически во всех науках о Земле. Приведем некоторые примеры исследовательских работ, которые используют DST для решения актуальных задач обработки изображений, полученных при дистанционном зондировании:

– классификация объектов и групп объектов на много- и гиперспектральных аэрокосмических изображениях [47];

– улучшение качества изображений, полученных в результате дистанционного зондирования, путем обнаружения изменений в изначально дефектных снимках [48].

**Строительство.** Еще одной важной областью применения методов принятия решений в условиях неопределенности контролируемых факторов, основанных на использовании DST, является строительство. Отметим, как примеры, следующие направления исследований и разработок:

– анализ и оценка состояния мостов [49];

– разработка методов расчета надежности грунтовых оснований зданий и сооружений [50];

– расчеты надежности железобетонной балки при ее эксплуатации [51];

– оценка состояния и анализ дефектов канализационных трубопроводов, соединений и люков [52].

## Заключение

Представленный обзор иллюстрирует активное развитие теории Демпстера-Шафера, востребованность ее результатов для решения практических задач. Это касается довольно большого количества самых разных научных и технологических областей. Дополнительно к представленным в обзоре отметим и некоторые другие актуальные направления:

– анализ и управление рисками, например, управление рисками при планировании производства новых продуктов;

– решение задач классификации интервальных данных, совместное использование различных классификаторов;

– системы геолокации: навигационные системы с беспроводными датчиками, системы обеспечения безопасности мореплавания (предотвращение инцидентов);

– решение задач оптимизации в условиях неопределенности параметров;

– гуманитарные области, такие как медицина или социология: методика обнаружения вируса Эболы, анализ результатов магнитно-резонансная томография, моделирование поведенческих норм в социологических исследованиях.

Есть основания полагать, что развитие отмеченных в обзоре тенденций в перспективе не станет менее интенсивным. Следовательно, дальнейшее совершенствование теории Демпстера-Шафера, особенно в части расширения граничных условий ее применимости, останется актуальной и востребованной задачей и очевидным стимулом для дальнейших исследований и разработок.

## Литература

## Current Trends and Applications of Dempster-Shafer Theory (Review)

V.K. Ivanov[I], N.V. Vinogradova[I], B.V. Palyukh[I], A.N. Sotnikov[II]

[I] Tver State Technical University, Tver, Russia
[II] Joint Supercomputer Center of RAS, Moscow, Russia

The article provides a review of the publications on the current trends and developments in Dempster-Shafer theory and its different applications in science, engineering, and technologies. The review took account of the following provisions with a focus on some specific aspects of the theory. Firstly, the article considers the research directions whose results are known not only in scientific and academic community but understood by a wide circle of potential designers and developers of advanced engineering solutions and technologies. Secondly, the article shows the theory applications in some important areas of human activity such as manufacturing systems, diagnostics of technological processes, materials and products, building and construction, product quality control, economic and social systems. The particular attention is paid to the current state of research in the domains under consideration and, thus, the papers published, as a rule, in recent years and presenting the achievements of modern research on Dempster-Shafer theory and its application are selected and analyzed.

**Keywords:** probability, evidence combination, conflict simulation, decision-making, the theory of evidence, a plausibility degree, Dempster-Shafer theory, a belief function, uncertainty accounting.